\documentclass[preprint,12pt]{elsarticle}

\usepackage{graphicx}
\usepackage{lineno}
\usepackage{amssymb}
\usepackage{array,multirow}
\usepackage{xcolor}
\usepackage{comment}
\usepackage{tabularx}
\usepackage{graphicx}
\usepackage{amssymb}
\usepackage{titlesec}
\usepackage{array}
\newcolumntype{P}[1]{>{\centering\arraybackslash}p{#1}}
\usepackage{booktabs}
\usepackage{xspace}
\usepackage{url}
\usepackage{geometry}
\usepackage{pdflscape}
\usepackage{tikz}
\usetikzlibrary{shapes.geometric, arrows}
\usepackage{lineno}
\usepackage[latin1]{inputenc}
\usepackage{multirow}
\usepackage{xcolor}
\usetikzlibrary{shapes,arrows}

\usepackage{lineno}

\usepackage{lipsum}
\makeatletter
\def\ps@pprintTitle{%
 \let\@oddhead\@empty
 \let\@evenhead\@empty
 \def\@oddfoot{}%
 \let\@evenfoot\@oddfoot}
\makeatother

\begin{document}

\begin{frontmatter}


\title{Multiclass Burn Wound Image Classification Using Deep Convolutional Neural Networks}



\author{Behrouz Rostami\textsuperscript{1}}
\author{ Jeffrey Niezgoda\textsuperscript{2}}
\author{ Sandeep Gopalakrishnan\textsuperscript{3}}
\author{ Zeyun Yu\textsuperscript{1,4}}

\address{\textsuperscript{1}Department of Electrical Engineering, University of Wisconsin-Milwaukee, Milwaukee, WI, United States}
\address{\textsuperscript{2}AZH Wound and Vascular Center, Milwaukee, WI, United States}

\address{\textsuperscript{3}College of Nursing, University of Wisconsin-Milwaukee, Milwaukee, WI, United States}

\address{\textsuperscript{4}Department of Computer Science, University of Wisconsin-Milwaukee, Milwaukee, WI, United States}

\address{\textsuperscript{*}Corresponding authors: \\Zeyun Yu, Email: yuz@uwm.edu\\
Sandeep Gopalakrishnan, Email: sandeep@uwm.edu
}

\begin{abstract}
 Millions of people are affected by acute and chronic wounds yearly across the world. Continuous wound monitoring is important for wound specialists to allow more accurate diagnosis and optimization of management protocols. Machine Learning-based classification approaches provide optimal care strategies resulting in more reliable outcomes, cost savings, healing time reduction, and improved patient satisfaction. 
In this study, we use a deep learning-based method to classify burn wound images into two or three different categories based on the wound conditions. A pre-trained deep convolutional neural network, AlexNet, is fine-tuned using a burn wound image dataset and utilized as the classifier. The classifier's performance is evaluated using classification metrics such as accuracy, precision, and recall as well as confusion matrix. A comparison with previous works that used the same dataset showed that our designed classifier improved the classification accuracy by more than 8\%. 
\end{abstract}


\end{frontmatter}


\section{Introduction}
\label{S:1}
Wound care and treatment accumulate huge costs to the healthcare frameworks worldwide and influence millions of people yearly. Just in the U.S., billions of dollars are spent annually to manage the therapeutic expenses~\cite{demidova2012acute,sen2009human}. On the other hand, wound classification and continuous precise monitoring of the wound healing progress will help clinicians to assess the efficacy of treatment and to identify early signs of stagnation or deterioration. Therefore, having an efficient wound classifier with a reliable classification performance  will result in saving time and money.
\\
As a potential solution for a wide range of human life problems, artificial intelligence (AI) has been used in a variety of domains such as manufacturing, policing, marketing, and healthcare within the past few decades~\cite{jordan2015machine}. Specifically, we have seen significant influences of AI and its subbranches like Machine Learning (ML) in healthcare~\cite{esteva2019guide, jiang2017artificial}. Radiology, ophthalmology, immunology, genetics, and wound care are just a few examples of ML used in these areas.  ~\cite{lakhani2018machine,figgett2019machine,andreatta2017machine,bari2017machine,rahman2018machine,collier2019lotus,yu2018artificial}. ML assists physicians in various ways such as disease diagnosis, prognosis, and treatment planning.~\cite{ngiam2019big}.\par
Moreover, a newer ML subbranch called Deep Learning (DL) has also been utilized widely in various fields and generated considerable outcomes, especially in medical image analysis domain. The increasing growth of the data complexity and volume in the medical imaging realm has resulted in broadly utilization of deep learning in the area~\cite{jiang2017artificial}. Detection of organs and body parts in MR or CT images, cell detection in histopathological images, and computer-aided detection and diagnosis are health care examples in which researchers used DL and its derivatives such as Deep Convolutional Neural Networks (DCNN)\cite{shen2017deep}. DCNNs are one of the most common deep learning-based models with a varying number of hidden layers. They are different from the classical neural networks that have a limited number of layers between the network's input and output~\cite{jiang2017artificial}. In a DCNN, convolution is the main mathematical operation to process the network's input~\cite{rostami2019survey}.\\
To address the recent wound image analysis challenges, many DCNN-based methods were proposed for the processing of wound images, including classification~\cite{wang2015unified,li2018composite,rajathi2019varicose,goyal2019skin,veredas2015wound,abubakar2019discrimination,zahia2018tissue,zhao2019fine}. 
In these studies, different types of wounds including diabetic, pressure, and burn ulcers were studied~\cite{goyal2019skin,veredas2015wound,abubakar2019discrimination,zahia2018tissue,zhao2019fine}.
However, only a limited number of publications are available for burn wound image classification using DCNNs. Most of these prior studies classify burn images into only two categories. In this manuscript, we propose an end-to-end deep learning-based method for classifying burn wound images into two and three classes. A comparison with previously published work which used the same dataset as we utilized, is provided to evaluate the performance of our proposed method. \par
The rest of this paper has been organized as follows: Subsection~\ref{S:2} reviews recently published papers in the field of burn wound image classification. In Section~\ref{S:3} we describe the dataset utilized in this study and discuss the method for designing the classifier. Section~\ref{S:4} presents the experimental results and discussion. Finally, we conclude the paper in Section~\ref{S:6}.

\subsection{Related works}
\label{S:2}
\begin{figure}[h!]

\centering\includegraphics[width=0.9\linewidth]{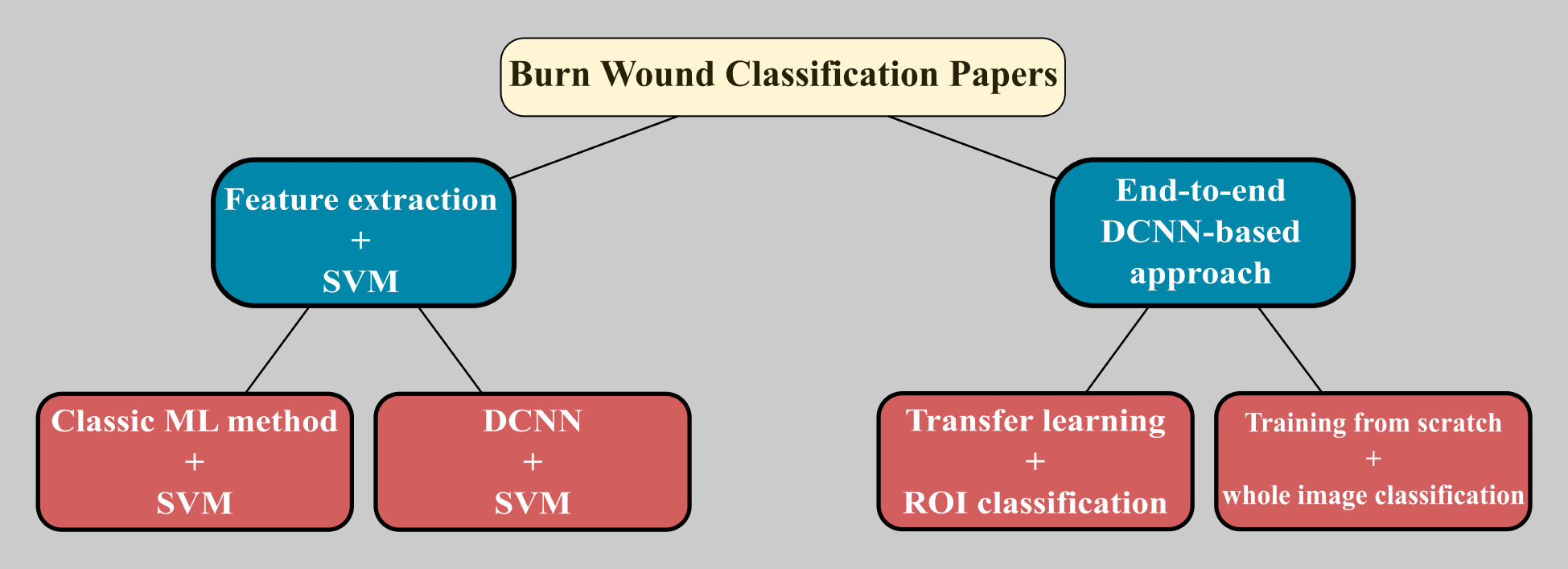}

\caption{Organization chart for the burn wound classification papers.}
\label{fig:org}
\end{figure}

In this subsection, some of the recent publications in burn wound image classification have been reviewed and categorized under two main subbranches: feature extraction methods along with an SVM, and end-to-end DCNN-based approaches. Figure~\ref{fig:org} displays the complete organization chart for the studied papers.\par

\subsubsection{End-to-end DCNN-based approaches}
Chauhan et al. used deep learning approaches to classify burn images based on the part of the body recognized in the image~\cite{chauhan2018using}. The proposed method used ResNet-50 architecture to categorize the input burn images into the face, hand, back, and inner arm classes. The utilized dataset included 109 burn images collected from the web as well as 4981 non-burnt images obtained from some available datasets. Two dependent (which used leave-one-out cross-validation) and independent (which used an independent deep learning model) approaches were tested in this research. The second strategy generated a better classification performance with an accuracy value of higher than 93\%. \par 
In another article, Cirillo et al. applied deep convolutional neural networks on burn images to predict the wound depth~\cite{cirillo2019time}. The goal was to classify the burn images into four classes: deep partial-thickness and full-thickness depth, intermediate to deep partial thickness, superficial to intermediate partial thickness, and the superficial partial thickness. A total of 23 burn images collected by a hospital in Sweden were used for training the models. 676 extracted ROIs from six classes (four burn-depth classes as well as normal skin and background categories) were augmented to overcome the overfitting problem. Four pre-trained deep architectures including VGG-16, GoogleNet, ResNet-50, and ResNet-101 were tested as classifiers. The ResNet-101 generated the best classification performance with the average classification accuracy of 91\%. Moreover, the authors claimed that the data augmentation improved the classification accuracy.  

\subsubsection{Feature extraction + SVM}
 Yadav et al. suggested a machine learning-based approach for binary classification of burn wound images~\cite{8766148}. A traditional color-based feature extraction method was used in company with a support vector machine to classify the images into two classes (grafted and non-graft). The utilized dataset contained 94 images from three burn types including full-thickness, deep dermal, and superficial dermal. The first two types constituted the "grafted" class while the "non-graft" category consisted of the superficial dermal images. The reported classification accuracy was 82.43\%.\par
 Abubakar et al. used DCNNs for binary classification of pressure and burn wound images~\cite{abubakar2019can}. Different deep architectures including VGG-face, ResNet101, and ResNet152 were applied for feature extraction followed by an SVM to classify the images into burn or pressure categories. The dataset was collected from the internet as well as a hospital source and contained 29 pressure and 31 burn wound images. Cropping, rotation, and flipping transformations were utilized for data augmentation. Also, several binary (burn or pressure) and 3-class (burn, pressure, or normal skin) classification experiments were conducted. For both classification problems, the best performance was related to ResNet152 which resulted in classification accuracy of 99.9\%.In a second study, Abubakar et al. proposed a classification method based on deep learning to classify burn wound images into Caucasian and African patient categories~\cite{abubakar2019noninvasive}. Three pre-trained deep convolutional neural networks including VGG-16, VGG-19, and VGG-Face, were used for feature extraction from the input images. In the next step, an SVM classified the extracted features into one of the healthy or burn classes. For each of the deep architectures, three different datasets including African patients, Caucasian patients, and a combination of them were utilized to train the SVM. The intent was to study how the combination of the images from different skin colors affects the classifier's performance. The dataset consisted of 32 Caucasian and 60 African cases. During the patch generation step, 1360 and 700 patches were extracted from the two groups, respectively. The authors mentioned that the classification accuracy was higher for Caucasian patients in comparison to the African patients or to the hybrid group. The combination of VGG-16 and SVM was reported as the best classification strategy with an accuracy of 99.286\% for Caucasian patients, 98.869\% for African patients, and 98.750\% for the hybrid dataset. ResNet101 architecture followed by an SVM classifier was used in another research by Abubakar et al. to classify burn wound images into one of the two classes, burn or normal~\cite{abubakar2019discrimination}. A pre-trained ResNet101 architecture was utilized as a feature extractor and then an SVM was applied for classification. The dataset was collected from the internet and after augmentation included 1360 images. The reported accuracy and precision values were 99.49\% and 99.56\%, respectively. Also, the outcomes were compared with LeNet's performance in a similar article which presented 81.81\% for the precision metric.

Table~\ref{tab:works} summarizes the reviewed papers. Based on this literature review, only a few articles studied the burn wound image classification problem and most of them only discussed the binary classification task. Additionally, it is important to note that only a limited number of researches used an automatic end-to-end deep learning-based method for classification. Instead, at the final stage they utilized a traditional ML tool as the classifier. To address this research gap, we propose an end-to-end DCNN-based approach to perform binary and 3-class classification of burn wound images. 


\begin{table}[]
\centering
\small
\caption{Summary of burn wound image classification works.}
\label{tab:works}
\begin{tabular}{|m{3.5cm}|m{4cm}|m{3.5cm}|m{3.5cm}|}
\hline
\textbf{Work/Research} & \textbf{Classification} & \textbf{Methods} & \textbf{Dataset} \\ \hline

Yadav et al.~\cite{8766148}  &      Binary classification of burn images into wounds that needs graft and the non-graft wounds   &  Support Vector Machine (SVM). &    Burns BIP\_US Database.  \\ \hline

Abubakar et al.~\cite{abubakar2019can} & Binary Classification of wound images into burn wounds or pressure wounds.    &   VGG-face, ResNet101, and ResNet15 networks with SVM classifier  & A dataset with 29 pressure wound images and 31 burn wound images.\\ \hline

 Abubakar et al.~\cite{abubakar2019noninvasive}    &    Binary Classification of input images into healthy or burn wounds.                        &        VGG-16, VGG-19, and VGG-Face with SVM classifier              &        A dataset with 32 wound images from Caucasians and 60 wound images from Africans      \\ \hline

 Abubakar et al.~\cite{abubakar2019discrimination}.    &    Binary Classification of input images into burn skin or normal skin.                        &        ResNet101  architecture  and  SVM  classifier               &        An unknown number of burn wound images collected from the internet        \\ \hline
 
 Chauhan et al.~\cite{chauhan2018using}.    &        Classifying the body part of burn images into four classes: back, face, hand, and inner forearm                    &              ResNet-50 architecture         &       109 burn images collected from the web as well as 4981 non-burnt image         \\ \hline
 
 Cirillo et al.~\cite{cirillo2019time}.    &      Classifying the burn images into four classes: deep partial-thickness and full-thickness, intermediate to deep partial thickness, superficial to intermediate partial thickness, and superficial partial thickness                      &    VGG-16,  GoogleNet,  ResNet-50,  and  ResNet-101                   &       23 burn images collected by a hospital in Sweden         \\ \hline

\end{tabular}
\end{table}

\section{Materials and Methods}
\label{S:3}

\subsection{Dataset}
\begin{figure}[htp]
\centering
\includegraphics[width=.2\textwidth,height=.2\textwidth]{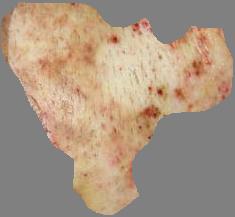}\quad
\includegraphics[width=.2\textwidth,height=.2\textwidth]{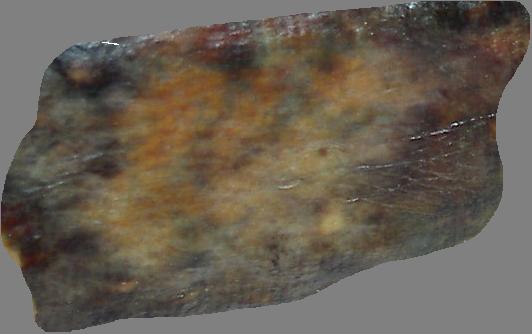}\quad
\includegraphics[width=.2\textwidth,height=.2\textwidth]{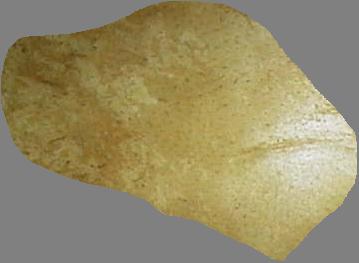}\quad
\includegraphics[width=.2\textwidth,height=.2\textwidth]{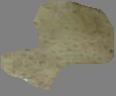}

\medskip

\includegraphics[width=.2\textwidth,height=.2\textwidth]{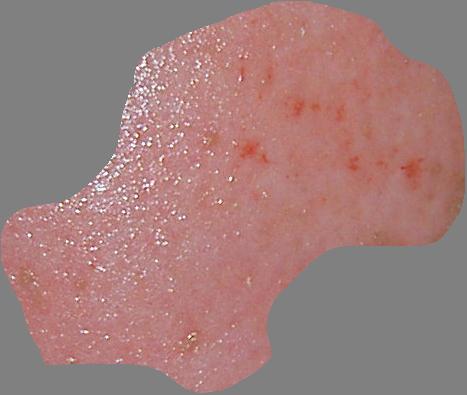}\quad
\includegraphics[width=.2\textwidth,height=.2\textwidth]{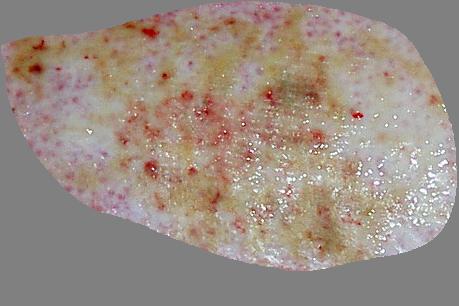}\quad
\includegraphics[width=.2\textwidth,height=.2\textwidth]{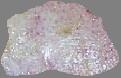}\quad
\includegraphics[width=.2\textwidth,height=.2\textwidth]{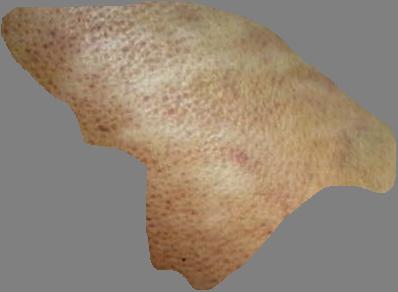}

\medskip

\includegraphics[width=.2\textwidth,height=.2\textwidth]{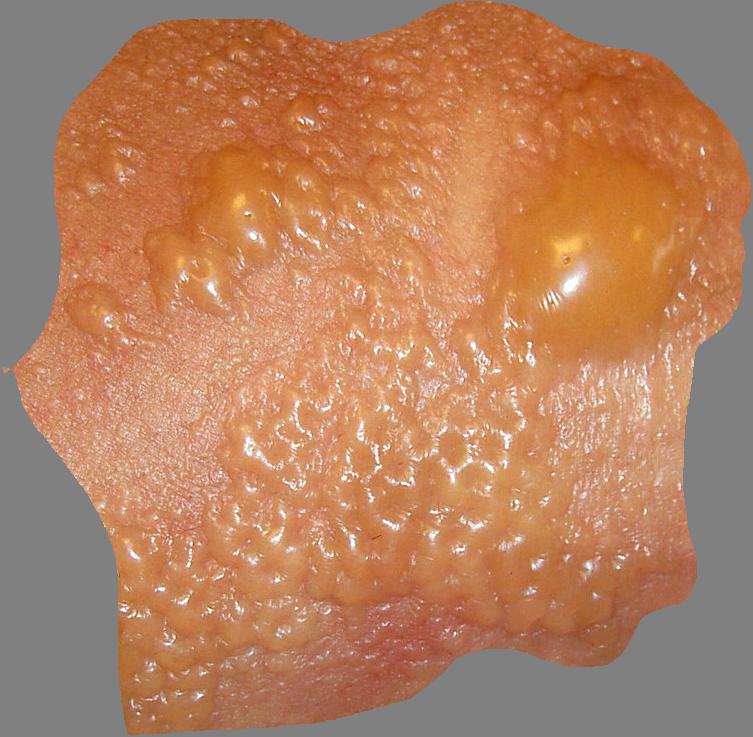}\quad
\includegraphics[width=.2\textwidth,height=.2\textwidth]{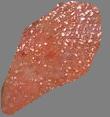}\quad
\includegraphics[width=.2\textwidth,height=.2\textwidth]{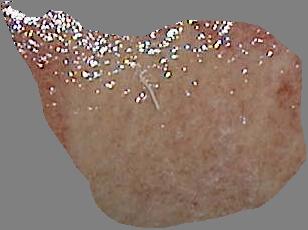}\quad
\includegraphics[width=.2\textwidth,height=.2\textwidth]{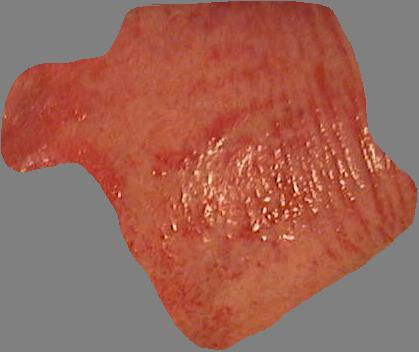}

\caption{BIP\_US database sample images. The rows from top to bottom display full-thickness, deep dermal, and superficial dermal samples, respectively.}
\label{fig:samples}
\end{figure}

In this research, the same dataset that was utilized in a prior investigation~\cite{8766148} was used for comparing our results with those previously reported. As mentioned in Section~\ref{S:2}, this dataset, BIP\_US, contains 94 images from three burn wound types including full-thickness, deep dermal, and superficial dermal~\cite{BIP}. There are 20, 32, and 42 samples in each class, respectively. The images have jpg and bmp formats and they are in different sizes. Figure~\ref{fig:samples} shows some dataset samples.


\subsection{Methods}
\subsubsection{Preprocessing}

\textbf{Data splitting:}~In the 3-class classification problem we split the data into train, validation, and test sets with 76, 9, and 9 images, respectively. For the binary classification experiment, since the goal was to compare our method with the approach used in~\cite{8766148}, we followed the same data splitting strategy and put 74 images into the test set and the rest of the data samples into the train and validation sets.

\textbf{Data Augmentation:}~The training data were augmented by generating 16 images from each image using transformations like rotating, flipping, cropping, and mirroring. Therefore, for the 3-class classification experiment, the number of training samples in the classes deep dermal, full-thickness, and superficial dermal increased to 416, 224, and 576 after augmentation. For the binary classification case, we ended up with 128 images in the non-grafted class and 144 images in the grafted category.

\subsubsection{Training the DCNN using transfer learning}
Due to the limited number of the images in the dataset and the complexity level of the classification problem, the AlexNet architecture was used in this research. AlexNet is a deep convolutional neural network proposed in 2012 as the winner of ILSVRC which outperformed all traditional machine learning methods~\cite{krizhevsky2012imagenet,alom2018history}. AlexNet consists of 8 layers, including 3 convolution and 2 fully connected layers, with 60 million parameters. This network is one of the most popular deep architectures in computer vision applications such as classification tasks~\cite{alom2018history}. We trained AlexNet using the transfer learning strategy. By using this method, an AlexNet that was pre-trained on ImageNet~\cite{ImageNet}, was fine-tuned using the BIP\_US dataset samples. ImageNet is a huge dataset that includes more than 14 million general images. All the experiments in this study were implemented in version R2020a of MATLAB software. We used an Intel(R) Core (TM) i7-8565U CPU @1.80GHz 1.99 GHz and NVIDIA GEFORCE MX 150 GPU with 2GB of memory to run the experiments.



\section{Results and Discussion}
\label{S:4}
\begin{figure}[h!]

\centering\includegraphics[width=0.7\linewidth]{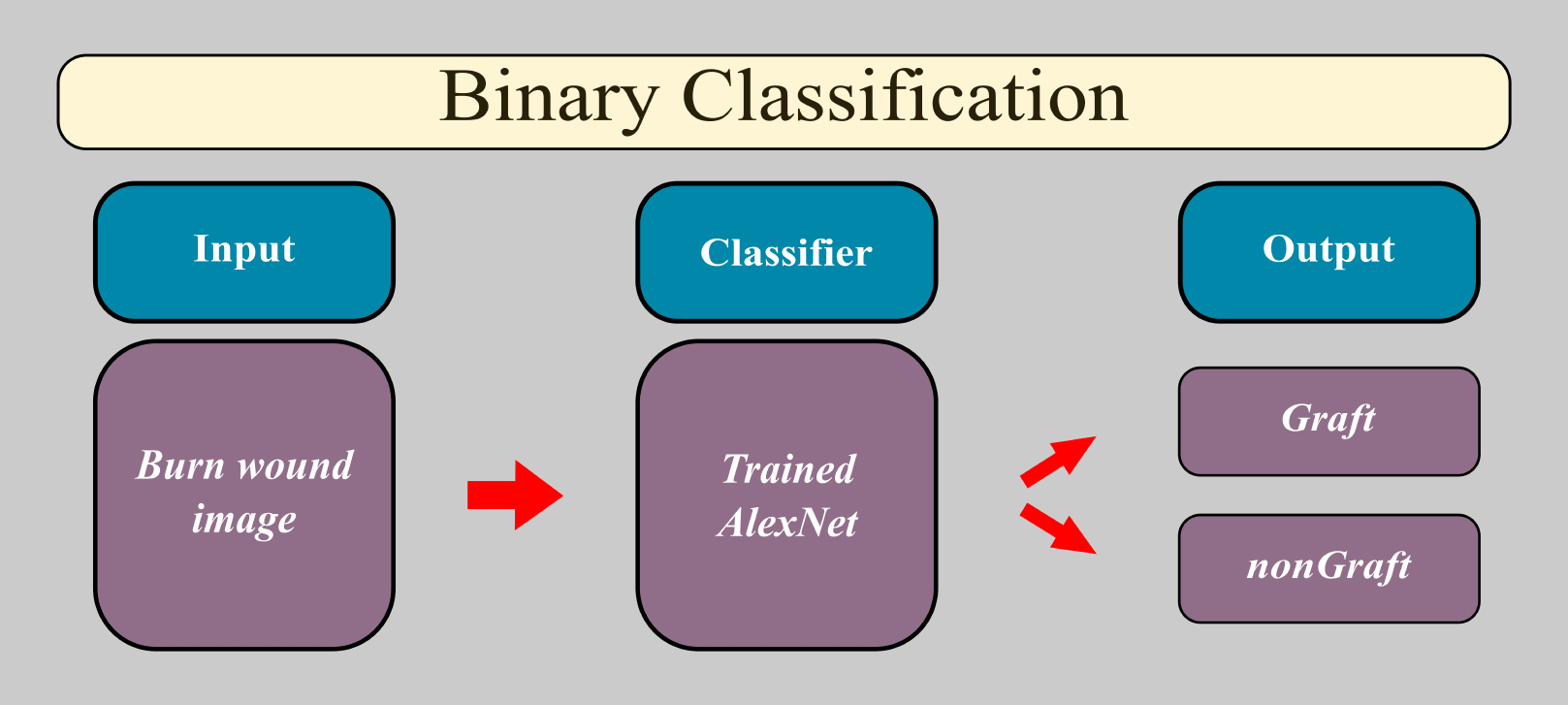}

\caption{Binary classification process.}
\label{fig:binary}
\end{figure}

\begin{figure}[h!]

\centering\includegraphics[width=0.7\linewidth]{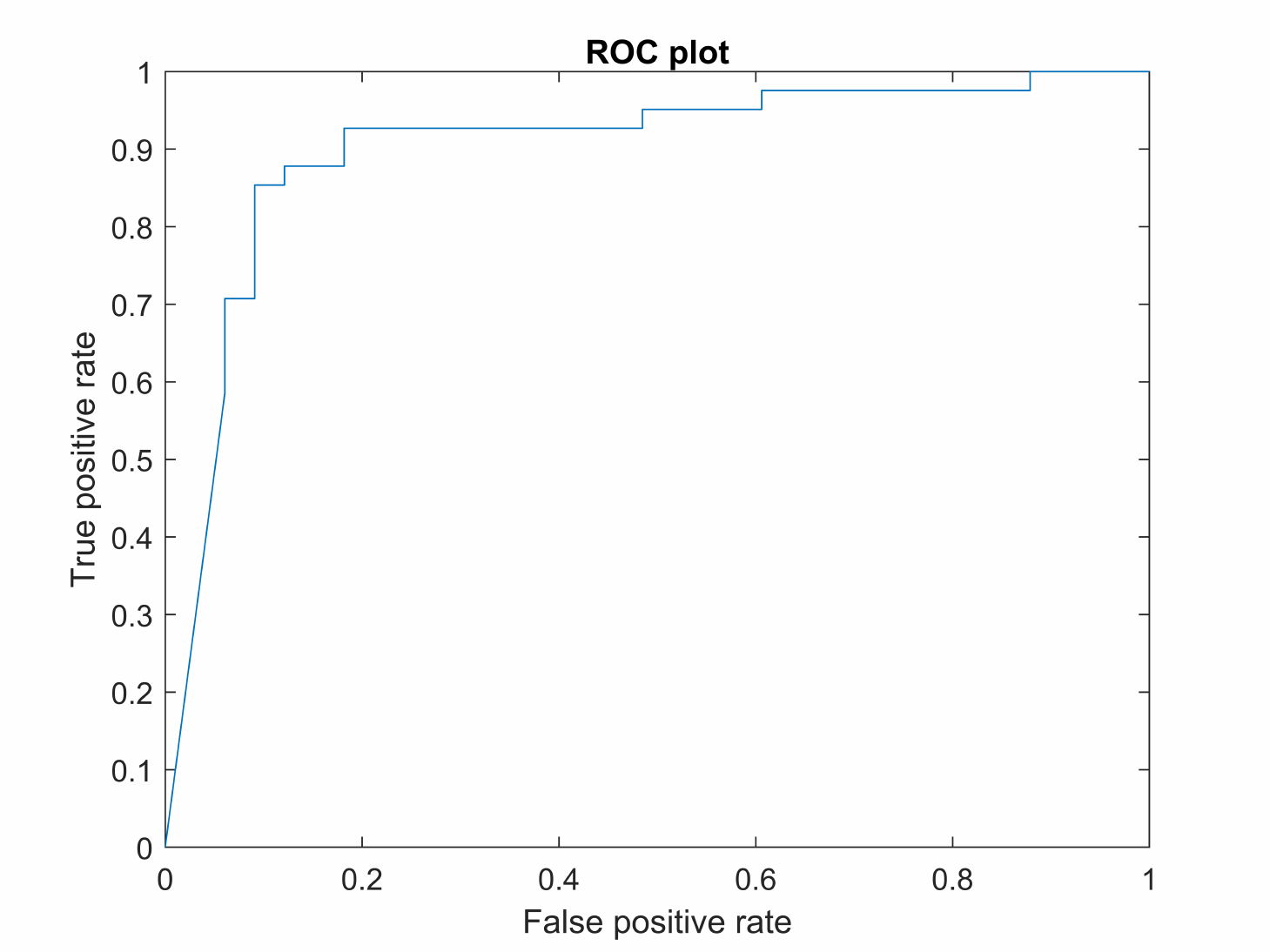}

\caption{ROC plot for binary classification.}
\label{fig:roc}
\end{figure}

ML and DL have been used widely in the literature to solve healthcare problems including wound care challenges. Different DL-based methods were proposed for analysis of wound images recently. However, the number of studies that cover the burn wound challenges including wound classification is limited. Also, in most cases the researchers discussed only the binary classification problem of burn wound images. In this article, we proposed an end-to-end DCNN-based strategy to classify the burn wound images into two and three categories. For the binary classification problem, we implemented the same experiment that was described in~\cite{8766148} for planned comparison. In this problem, the goal was to classify an input image into one of the two classes: the first class includes the burn wound images that required a grafting, and the second class included the non-grafted wound samples. Figure~\ref{fig:binary} shows the classification process in this experiment. The obtained results for this experiment have been summarized in Table~\ref{tab:binary} and Figure~\ref{fig:roc}. We selected 0.0001 as the learning rate and trained the network for 10 epochs with the mini-batch size of 64. As we observe in Table~\ref{tab:binary}, the test accuracy is 90.5\%. Also, the recall, precision, F1-score, and AUC values are 87.9\%, 90.6\%, 0.8922, and 0.913, respectively.

\begin{table}[h!]
\caption{Binary classification results}
\label{tab:binary}
\small
\centering
\begin{tabular}{|c|c|c|c|c|c|}
\hline
\textbf{Classification} &
\textbf{Accuracy (\%)} & \textbf{Precision (\%)} & \textbf{Recall (\%)} & \textbf{F1-score} &
\textbf{AUC} \\ \hline
          Graft/nonGraft    &     90.5       &       90.6     & 87.9 & 0.8922 & 0.913  \\ \hline
           
\end{tabular}
\end{table}


\begin{figure}[h!]

\centering\includegraphics[width=0.7\linewidth]{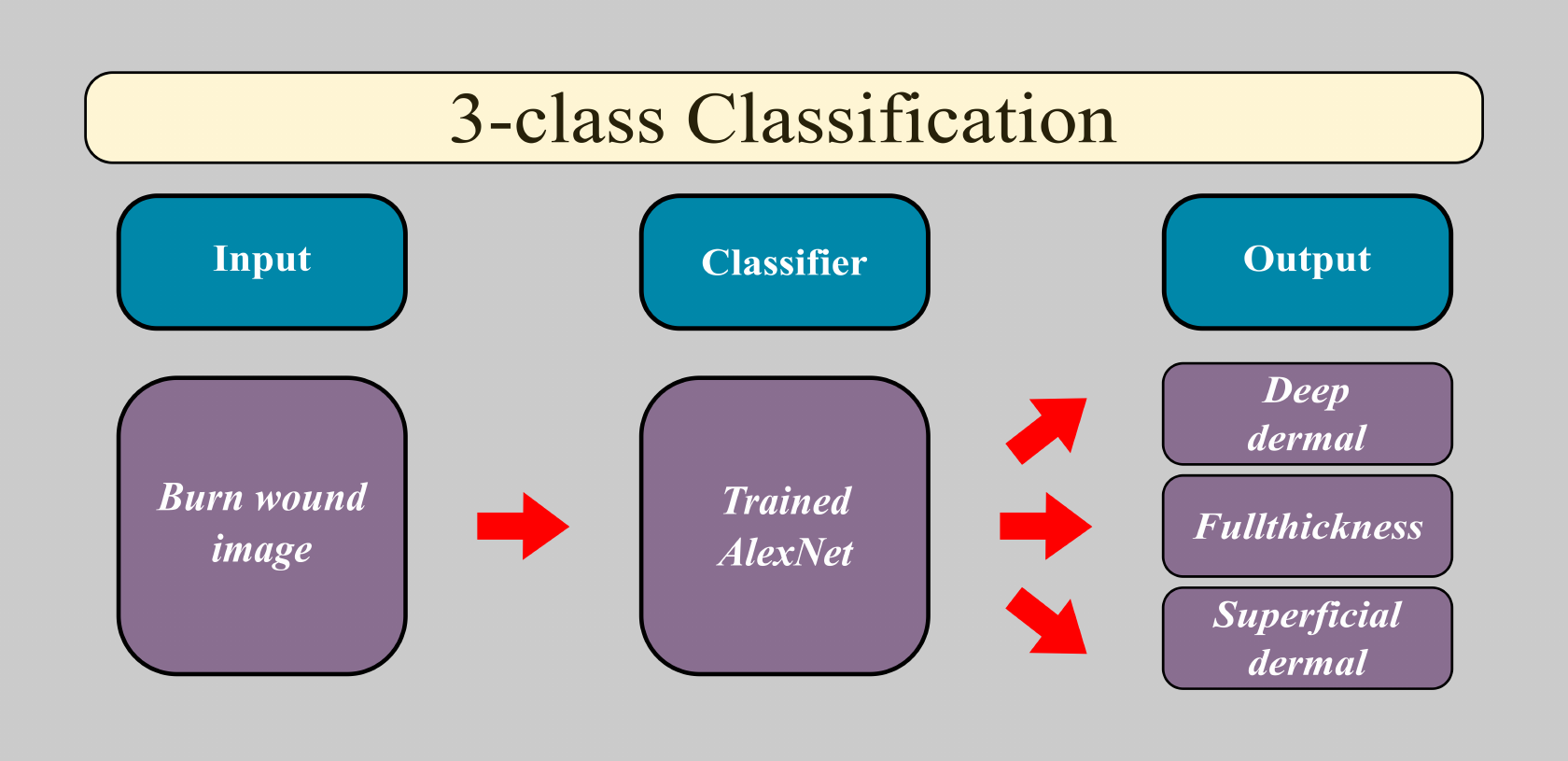}

\caption{3-class classification process.}
\label{fig:threeclassification}
\end{figure}

\begin{figure}[h!]

\centering\includegraphics[width=0.7\linewidth]{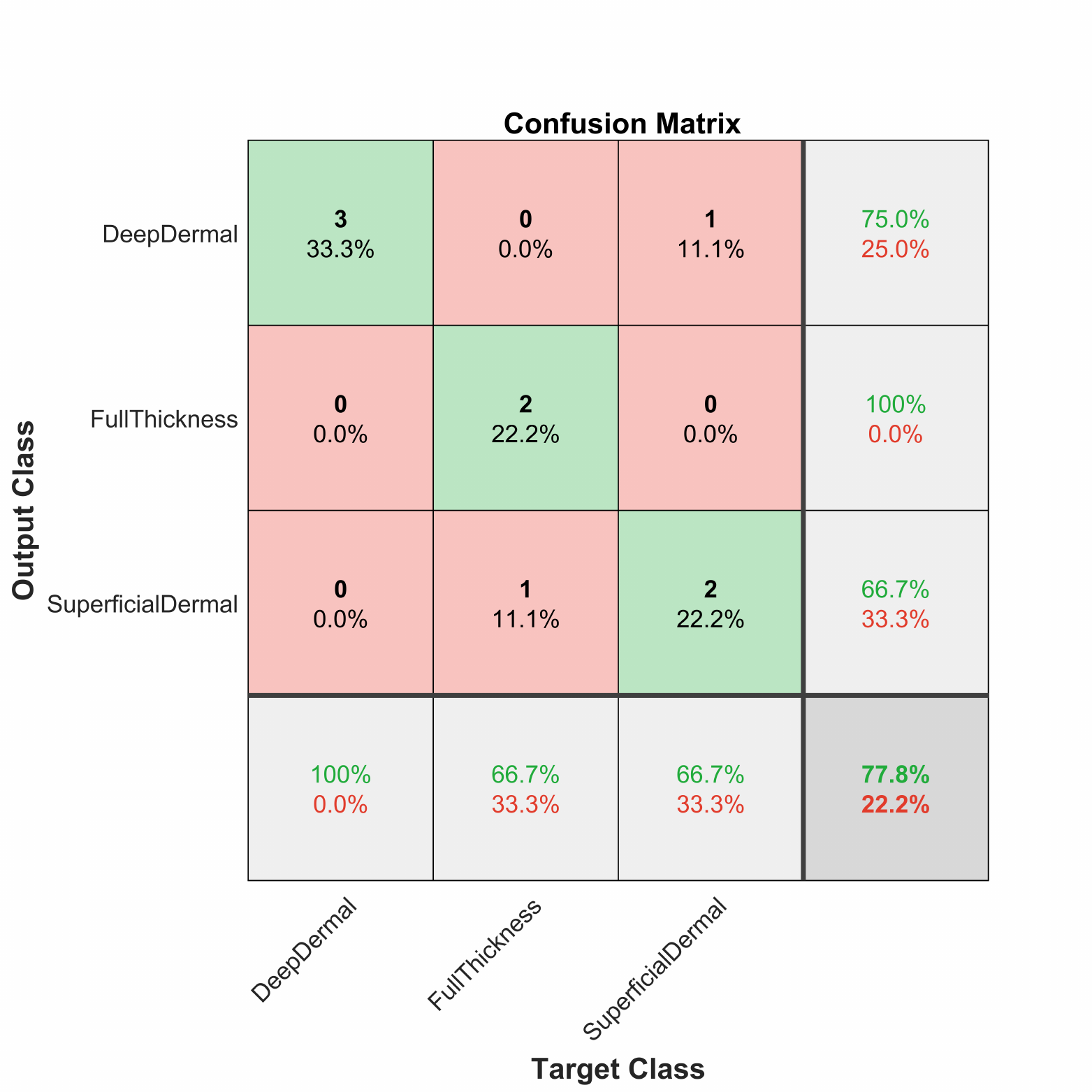}

\caption{Confusion matrix for 3-class classification problem.}
\label{fig:3class}
\end{figure}

For the 3-class classification experiment, the intent was to classify the input image into one of the three classes: full-thickness, deep dermal, and superficial dermal. Figure~\ref{fig:threeclassification} displays the classification process for this experiment. We used 1e-6 as the learning rate value and trained the network for 5 epochs with the mini-batch size of 10. From the confusion matrix displayed in Figure~\ref{fig:3class}, we observe that the test accuracy is 77.8\%. \par
Regarding the results discussed above, we find that by increasing the number of classes from two to three, the classification accuracy decreased considerably. The justification is that by increasing the number of classes the number of network parameters will grow. It means that the network needs more data for training the new parameters, otherwise the classifier's performance will drop. Based on the reported results in~\cite{8766148} on binary classification, the authors obtained an accuracy value of 82.43\% along with the precision, recall, and F1-score amount of 0.82, 0.88, and 0.85, respectively. As a conclusion, our designed classifier improved the binary classification accuracy by more than 8\%. By considering all the reported metrics, we claim that our classifier generated better performance for the binary classification problem. For the 3-class classification case, the confusion matrix shows that the deep dermal is the easiest wound class to be classified by the network. The other two classes show the same difficulty level of classification. Currently no existing work has been seen on 3-class classification of burn wound images for comparison with the proposed work.

\section{Conclusion}
Wound image classification is one of the most important stages during the treatment process and a precise classifier can help the clinicians to have more efficient diagnosis. Many machine learning and deep learning-based methods have been presented in recent years to design high-performance burn wound image classifiers. However, most of the previous studies discussed only the binary classification problem. To fill this gap, in this research we presented a deep learning-based method for end-to-end classification of burn wound images into two and three classes. We used a pre-trained AlexNet and fine-tuned it using a burn wound image dataset, BIP\_US. The results showed that our proposed approach can classify the burn wound images into two categories, grafted or non-grafted, satisfyingly. Also, our results display a considerable improvement over similar works in the literature. Moreover, despite having a very limited number of samples in the dataset, the proposed method provided a decent performance for classification of the burn wound samples into three categories: deep dermal, full-thickness, and superficial dermal. Both experiments in this study were limited by a small number of images and we expect that by using a larger dataset the results would be significantly improved. Our findings demonstrate that deep convolutional neural networks can be used successfully for burn wound image classification tasks or other similar clinical applications to improve the prognosis and treatment procedure.

\label{S:6}






\bibliographystyle{elsarticle-num-names}
\bibliography{sample.bib}







\end{document}